# Concrete Surface Crack Detection with Convolutional-based Deep Learning Models

Sara Shomal Zadeh[1], Sina Aalipour birgani[2], Meisam Khorshidi[3], Farhad Kooban[4]

[1]Department of Civil and Environmental Engineering, Lamar University, USA

[2]Master of Mechanical engineering -energy conversion, Sharif university of technology international campus kish island, Iran

[3]Department of Civil and Environmental Engineering, University of New Hampshire, Durham, NH 03824, USA.

[4]Department of Civil and Environmental Engineering, The Pennsylvania State University, University Park, PA 16802, USA

*Abstract:* **Effective crack detection is pivotal for the structural health monitoring and inspection of buildings. This task presents a formidable challenge to computer vision techniques due to the inherently subtle nature of cracks, which often exhibit low-level features that can be easily confounded with background textures, foreign objects, or irregularities in construction. Furthermore, the presence of issues like non-uniform lighting and construction irregularities poses significant hurdles for autonomous crack detection during building inspection and monitoring. Convolutional neural networks (CNNs) have emerged as a promising framework for crack detection, offering high levels of accuracy and precision. Additionally, the ability to adapt pretrained networks through transfer learning provides a valuable tool for users, eliminating the need for an in-depth understanding of algorithm intricacies. Nevertheless, it is imperative to acknowledge the limitations and considerations when deploying CNNs, particularly in contexts where the outcomes carry immense significance, such as crack detection in buildings. In this paper, our approach to surface crack detection involves the utilization of various deep learning models. Specifically, we employ fine-tuning techniques on pre-trained deep learning architectures: VGG19, ResNet50, Inception V3, and EfficientNetV2. These models are chosen for their established performance and versatility in image analysis tasks. We compare deep learning models using precision, recall, and F1 score.**

## 1. INTRODUCTION

In the realm of civil engineering, the detection and classification of surface cracks take center stage in the pursuit of maintaining the structural integrity of vast concrete infrastructure. The importance of these endeavors extends far beyond mere maintenance; they are the guardians of both human safety and the longevity of our architectural wonders. Large concrete structures, such as bridges, dams, and buildings, stand as enduring symbols of human ingenuity and progress [1]. However, the very forces that these grand constructs contend with – the relentless passage of time, environmental elements, and unforeseen events like natural disasters – conspire to inflict wear and tear upon them. It is in this delicate balance that surface cracks emerge, often imperceptible to the human eye, silently eroding the stability of these edifices. Their presence, if left unattended, can be the harbinger of potential disasters, with implications reaching far beyond structural integrity. The process of identifying, tracking, and ultimately mitigating these cracks assumes paramount significance in the realm of civil engineering. Early detection and precise classification of surface cracks can spell the difference between the preservation of an iconic structure and its potential descent into disrepair. The benefits reverberate on multiple fronts: public safety remains paramount, maintenance costs are reduced, the lifespan of constructions is prolonged, and the efficiency of infrastructure utilization is optimized. Within this intricate tapestry of challenges and opportunities, we delve into the domain of surface crack detection and classification. As we journey deeper, we unveil the transformative potential of advanced technologies and the prowess of deep learning models in contributing to the safety and sustainability of our architectural heritage. Our exploration is poised to shed light on the multifaceted facets of this field, where science, technology, and engineering converge to deliver invaluable benefits to civil engineering [1].

In recent years, an increasingly prominent pattern has emerged where machine learning and deep learning models are being utilized for image analysis across a range of applications, encompassing image recognition, classification, and segmentation in the realms of engineering and medicine [2-3]. However, in contrast to conventional and traditional methods employed in diverse civil engineering applications [4-9], deep neural networks, functioning as end-to-end learning models, exhibit the capability to automatically extract features from images. The fusion of deep learning and image analysis has ushered in a transformative era in the domain of structural assessment. Deep learning, particularly Convolutional Neural Networks (CNNs), has demonstrated its prowess in deciphering intricate visual patterns with unprecedented accuracy. In the context of surface crack detection, the impact of deep learning is profound. These powerful algorithms, trained on vast datasets of cracked surfaces, possess an inherent ability to discern subtle cracks from background textures, foreign objects, or environmental irregularities. The result is efficiency and precision that surpasses manual inspections. By automating the detection and classification of surface cracks, deep learning liberates civil engineers and inspectors from the constraints of subjectivity and protracted examination processes. Furthermore, it significantly reduces the risk of human error, ensuring that even the most inconspicuous cracks do not escape notice. Therefore, the reliability and safety of civil infrastructure are bolstered, maintenance costs are reduced, and the lifespans of structures are extended. The marriage of deep learning and image analysis is not merely a technological advancement; it is a safeguard for the preservation of our architectural heritage and the enhancement of public safety.

In this paper, we embark on a comprehensive exploration of surface crack detection and classification by harnessing the formidable capabilities of state-of-the-art deep learning models. Our arsenal includes the renowned VGG19 [10], the robust ResNet50 [11], the versatile Inception V3 [12], and the cutting-edge EfficientNetV2 [13]. These models, equipped with sophisticated architectures and a deep understanding of visual patterns, bring a wealth of expertise to the realm of crack detection. Their proven track records in diverse image analysis tasks position them as reliable candidates to navigate the intricate nuances of surface crack identification. By harnessing this ensemble of deep learning powerhouses, we aim to scrutinize their performance, assess their adaptability to varying conditions, and provide insights that will guide both newcomers and seasoned practitioners in the field of civil engineering toward effective and precise crack detection.

## 2. RELATED WORKS

Traditional approaches to detecting cracks typically involve a two-step process: feature extraction and classification. In the initial phase of feature extraction, image processing techniques are employed to capture relevant information regarding cracks (features) from images. These extracted features are then evaluated by different classifiers. Numerous studies in the field of crack detection have focused on these established machine learning methods [14-15]. However, if the extracted features fail to accurately represent the actual cracks, it can adversely affect the classifier's accuracy. Deep learning methods have significantly expanded the horizons and effectiveness of these conventional techniques, showcasing impressive performance in tackling the challenges of crack detection [16-20]. Deep convolutional neural network (CNN) models excel at extracting pertinent features from input data through multi-layer neural networks. Moreover, these models can be customized to enable crack localization using a sliding window approach, where a bounding box is applied to areas identified as cracks by the classifier. Building an efficient deep learning model requires careful selection of the appropriate network structure, loss function, and optimization algorithm.

Zhang et al. [21] introduced a six-layer convolutional neural network (CNN) architecture aimed at detecting cracks on pavement surfaces. They utilized datasets comprising 640,000, 160,000, and 200,000 images for training, validation, and testing, respectively. Wang et al. [22] proposed a CNN design that included three convolutional layers and two fully connected layers for the purpose of recognizing cracks in asphalt pavement. Their training data consisted of 640,000 image patches, with 120,000 patches allocated for testing. In another study [23], the authors combined a CNN architecture with sliding window techniques to identify and pinpoint cracks in concrete images using a dataset containing only 40 images. Their results indicated that training a network from the ground up was most effective when working with over 10,000 images.

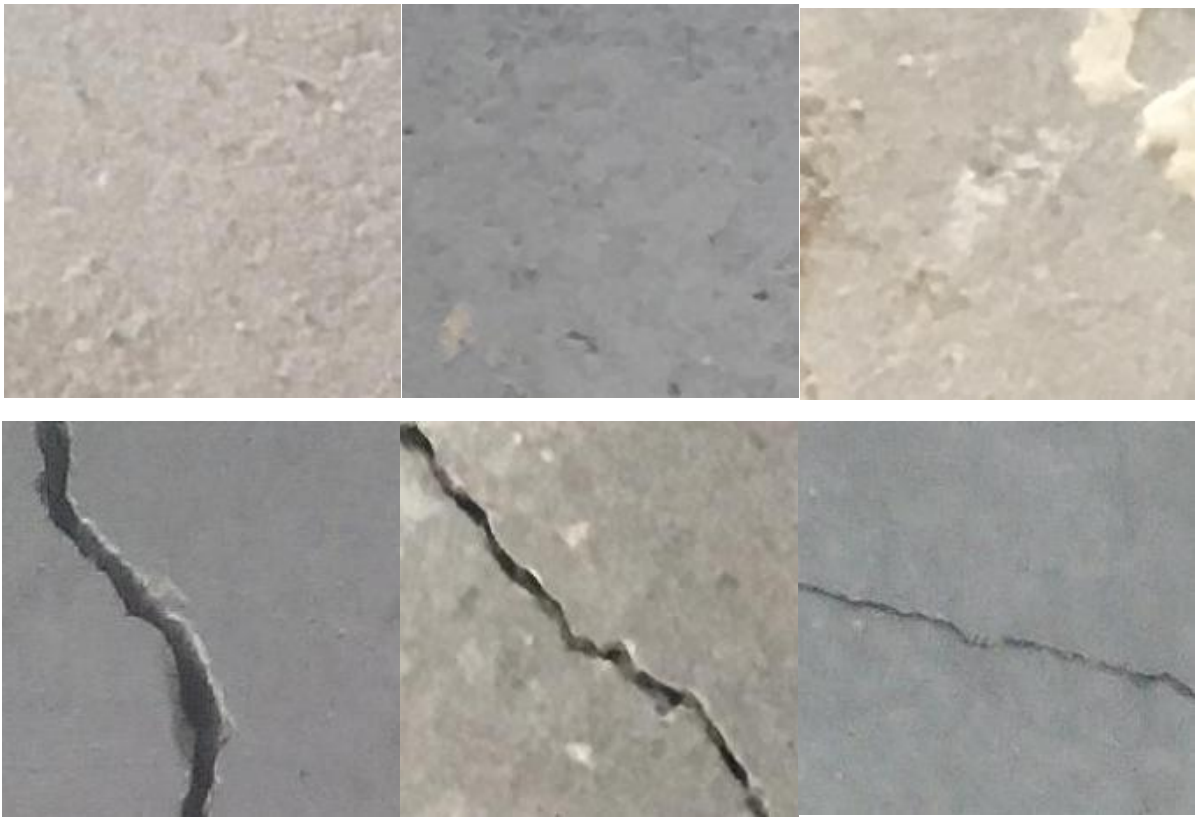

**Figure 1. Examples of images in dataset. first row: normal surface, second row: cracked surface**

Xu et al. [24] trained a 28-layer end-to-end CNN model tailored for detecting cracks in concrete bridge structures, employing a dataset of 6,069 images derived from a bridge. Their approach involved incorporating atrous spatial pyramid pooling (ASPP) to capture multiscale context information and utilizing depth-wise convolution to reduce network parameters. In yet another research endeavor [25], the authors introduced a CNN model for detecting cracks in pavement structures. They explored the impact of network depth and the positioning of images on model performance. Their findings suggested that increasing network depth improved performance but came at the cost of reduced generalization when altering the positioning of training and testing images. Fan et al. [26] proposed an efficient model for automatic pavement crack detection and measurement. This model utilized an ensemble of CNN models, and they applied a weighted average ensemble technique to compute the final crack probability by aggregating individual CNN model scores. In a different context [27], Tong et al. devised three distinct CNN architectures, each designed for specific functions like crack detection, localization, length measurement, and 3D reconstruction of hidden cracks in ground-penetrating radar (GPR) images. Yang et al. [28] introduced an innovative approach for the automatic detection and measurement of pixel-level cracked concrete structures using a deep learning approach known as a fully convolutional network (FCN). Their architecture encompassed both down-sampling, represented by conventional CNN layers, and up-sampling through deconvolutional layers. Zhang et al. [29] presented CrackNet, a CNN architecture comprising five layers (2 convolutional, 2 fully connected, and 1 output) designed for pixel-level crack detection on 3D asphalt surfaces. While CrackNet was effective in detecting pixel-level cracks, it suffered from long processing times and struggled to detect hairline cracks. To overcome these challenges, the authors introduced CrackNet II [30] to enhance processing speed, eliminate local noise, and improve hairline crack detection on 3D asphalt surfaces. These networks were developed from the ground up and required substantial training data and time. Training time can be minimized by fine-tuning pre-trained models for similar tasks.

Transfer learning models simplify the utilization of CNNs, eliminating the need for substantial computational resources or an in-depth understanding of CNN operation. These transfer learning models designed for image data intake encompass well-known options such as Google's VGGNet [10], Microsoft's ResNet [11], and Inception-V3 [12]. Gopalakrishnan et al. [31] harnessed a pre-trained VGG-16 architecture for pavement crack detection, demonstrating its effectiveness with a relatively small dataset comprising 760 training images, 84 validation images, and 212 test images. In comparison to previous CNN models, this transfer learning model exhibited reliability, speed, and ease of implementation. In a different context, Zhang and Cheng [32] employed an ImageNet-based pre-trained model to detect and repair cracks in pavement images. Their training dataset consisted of 30,000 crack patches and an equal number of non-crack patches, with 20,000 patches assigned to each category in the test set. In another study [33], the authors harnessed the VGGNet model with 2,000 labeled images (in a 4:1 training-to-test data ratio) to identify various forms of structural damage. Bang et al. [34] opted for a deep residual network that leveraged transfer learning for road crack detection, utilizing a dataset comprising 118 images. Meanwhile, Wilson and Diogo [35] undertook robust training of the VGG-16 model for the purpose of concrete crack detection, using 3,500 images captured by unmanned aerial vehicles (UAVs), with an 80/20 split between training and test data. Furthermore, Gopalakrishnan et al. [36] put forth an automated crack detection system based on the VGG-16 model, making use of images acquired via drones. Additionally, Feng and Zhang [37] modified the architecture of the Inception V3 model and employed transfer learning to detect structural damage in concrete water pipes.

## 3. METHODS

### 3.1 Surface Crack Detection Dataset

The dataset comprises images of concrete surfaces exhibiting cracks, sourced from various structures within the METU Campus [38-39]. It has been meticulously organized into two distinct categories: "negative" and "positive" crack images, facilitating image classification. Each class encompasses a total of 20,000 images, resulting in a comprehensive dataset of 40,000 images, all uniformly sized at 227 x 227 pixels and featuring RGB channels. This dataset emanates from a novel approach rooted in the work of Zhang et al. (2016), whereby 458 high-resolution images, originally spanning a substantial 4032x3024 pixel resolution, have been transformed into the dataset at hand. These high-resolution source images exhibit significant variability in terms of surface finish and illumination conditions, mirroring the real-world diversity encountered during structural inspections. It's pertinent to note that the dataset maintains its integrity without the application of data augmentation techniques, such as random rotation or flipping. This unaltered dataset faithfully represents the nuances of concrete surface cracks as encountered in their natural environment.

### 3.2 Proposed Model

In our pursuit of efficient and precise surface crack detection and classification, we have harnessed the formidable capabilities of four distinguished deep learning models, each with its unique strengths and architectural intricacies. These models, namely VGG19, ResNet50, Inception V3, and EfficientNetV2, have established themselves as stalwarts in the domain of image analysis and classification. For surface crack image classification, we fine-tuned pretrained deep learning models. We replaced their final classification layers with a custom layer for binary classification, enabling them to specialize in detecting surface cracks. The entire network, including the pretrained and new layers, was then trained to optimize parameters for classifying images into two categories: with or without visible cracks. This fine-tuning process is crucial for creating accurate and adaptable surface crack detection models.

### 3.2.1 Pre-trained models

The VGG-19 transfer learning model consists of 19 layers, including 16 convolutional layers and three fully connected layers. These layers utilize 3x3 filters with a stride and padding size of 1 pixel. The use of small kernel sizes helps keep the number of parameters in check and allows them to cover the entire image effectively. Within the VGG-19 model, a 2x2 max pooling operation with a stride of 2 is applied. This model achieved second place in classification and first place in positioning at the 2014 ILSVRC competition, boasting a total of 138 million parameters. VGGNet reinforced the idea that CNNs should have deep layer networks, enabling hierarchical interpretation of visual data.

ResNet50, a distinguished member of the ResNet (Residual Network) family, stands as a formidable deep learning model renowned for its architectural depth and structural ingenuity. Comprising a total of 50 layers, the model's architecture is characterized by the introduction of residual blocks, which revolutionize deep neural networks. These residual blocks contain skip connections that bypass one or more layers, enabling the model to train with exceptional depth without succumbing to the vanishing gradient problem. This profound structural innovation allows ResNet50 to capture intricate image features, even in the presence of complex and noisy data. In essence, the depth of this architecture not only accommodates increasingly complex visual data but also empowers the model to extract fine-grained details within the imagery. The network's 60.8 million parameters equip it to recognize the most nuanced variations in images, making it an invaluable asset in tasks such as surface crack detection and classification. The unique structural fortitude and architectural depth of ResNet50 make it a prime choice for addressing intricate challenges in image analysis and classification.

Inception V3, a product of Google's innovative deep learning endeavors, stands out as a multibranch architecture with unparalleled versatility. Comprising 48 layers in total, this model's architecture embraces the concept of factorized convolution, featuring filters of various dimensions to capture features at different scales. The Inception-V3 pre-trained model is a model with more than 20 million parameters and has been trained by one of the leading experts in the hardware industry. This model is composed of both symmetrical and asymmetrical building blocks, where each block includes a variety of components such as convolutional layers, average and max pooling layers, concatenation, dropout layers, and fully connected layers. Additionally, batch normalization is a common practice applied to the input of the activation layer in this model. Classification is carried out using the Softmax method.

EfficientNetV2 represents the zenith of computational efficiency and cutting-edge precision in deep learning models. Building upon the success of its predecessor, EfficientNet, this model optimizes the use of model parameters while delivering superior performance. EfficientNetV2 goes one step further than EfficientNet to increase training speed and parameter efficiency. This network is generated by using a combination of scaling (width, depth, resolution) and neural architecture search. The main goal is to optimize training speed and parameter efficiency. Also, this time the search space also included new convolutional blocks such as Fused-MBConv. With approximately 25.7 million parameters, the model harmonizes accuracy and efficiency, catering to scenarios where computational resources are at a premium. The synergy of computational efficiency and state-of-the-art precision positions EfficientNetV2 as a promising model for tasks demanding both speed and accuracy.

In summary, VGG19, ResNet50, Inception V3, and EfficientNetV2 each bring unique architectural characteristics and strengths to the table. VGG19 is known for its simplicity and uniform architecture. ResNet50's strength lies in its deep architecture and the use of residual blocks. Inception V3 is versatile, with a multi-branch architecture, and EfficientNetV2 excels in computational efficiency while maintaining accuracy. The choice among these models depends on the specific requirements of the task and the available computational resources.

### 3.2.2 Customized network

To perform the critical task of classifying surface crack images, a comprehensive approach was adopted. Pretrained deep learning models, including VGG19, ResNet50, Inception V3, and EfficientNetV2, were fine-tuned to harness their feature extraction capabilities specifically for crack detection. In the fine-tuning process, the final classification layers of these models were replaced by a custom fully connected layer designed for binary classification. This layer was adapted to detect the presence or absence of surface cracks, essentially transforming the pretrained models into specialized instruments for this task. Subsequently, the entire network, consisting of the pretrained layers and the added fully connected layer, underwent training. This training process allowed the models to learn from the provided dataset and optimize their parameters for the precise classification of surface crack images into two categories: those with visible cracks and those without. The fine-tuning approach is a pivotal step in the development of robust and accurate models for surface crack detection, enabling them to adapt to the specific nuances of the dataset and excel in binary classification.

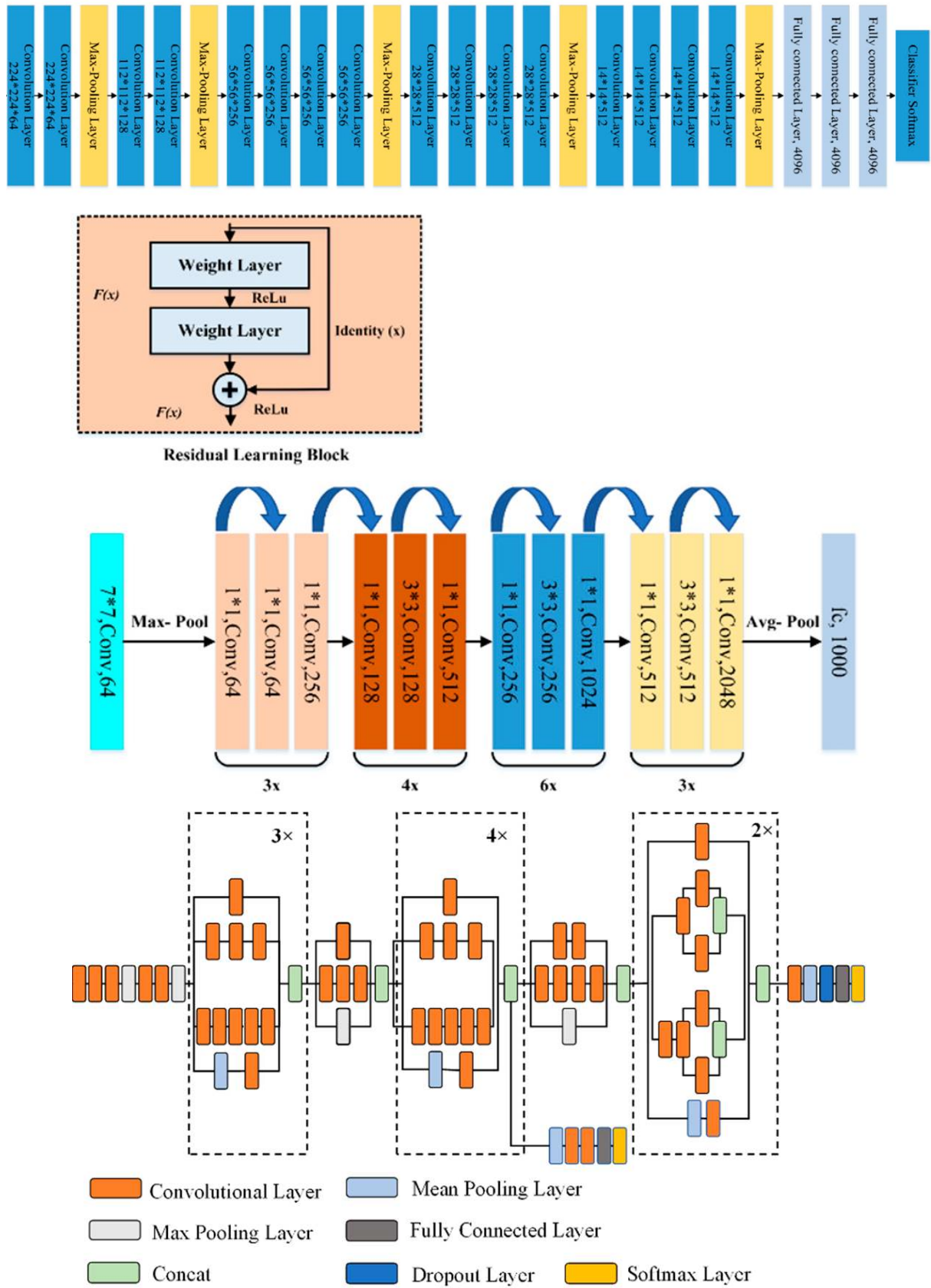

**Fig. 2. Top: The Architecture of VGG19, Middle: ResNet 50, Down: Inception V3 [1]**

# 4. RESULTS

## *4.1 Parameter Setting*

The deep learning model proposed in this study was developed using Keras. All experiments were carried out on a 4-core PC equipped with an i7-6700 processor running at 3.4GHz and 16GB of RAM and NVIDIA GeForce RTX 2090. Our approach involves employing a fixed margin of 0.2, implementing random sampling, and training the network over 500 epochs. We utilized a publicly available surface crack dataset, classifying it into two categories: "not cracked" and "cracked." As part of our preprocessing, we applied normalization and scaled the dimensions of all the images. In this research, we designated 80% of these images for training and validation, reserving the remaining 20% for the testing dataset.

## *4.2 Performance Metrics*

In our research, we utilized multiple evaluation metrics, including accuracy, precision, recall, and F1 score, to ensure a comprehensive comparison of our experimental outcomes. Accuracy is a measure of the ratio between correctly identified crack and non-crack patches and the total number of input patches. It involves TP (True Positive) and TN (True Negative) representing accurately identified crack and non-crack patches, and FP (False Positive) and FN (False Negative) representing patches incorrectly identified as crack or non-crack.

$$Accuracy = \frac{TP + TN}{TP + FP + TN + FN} \qquad (1)$$

$$Precision = \frac{TP}{TP + FP} \qquad (2)$$

$$Recall = \frac{TP}{TP + FN} \qquad (3)$$

$$F_{1_{score}} = 2 \times \frac{Precision \times Recall}{Precision + Recall} \qquad (4)$$

## *4.3 Results*

Table 2 and Tabel 3 provide a summary of our results, evaluating the performance of various models for concrete crack detection using different metrics. While Table 2 indicates the results of running the pre-trained models on Crack Dataset, the proposed customized model's performances have been shown in Table 3. Regarding Table 3, VGG19 achieved an accuracy of 92.2%, indicating the proportion of correctly classified patches, with high precision at 97.4%, reflecting its ability to minimize false positives. The recall, at 88.3%, signifies its capacity to identify actual crack patches, resulting in an F1 score of 92.9, which balances precision and recall. ResNet50 exhibited remarkable performance with an accuracy of 99.4% and high precision at 98.8%, signifying its proficiency in minimizing false positives. Its recall of 99.9% demonstrates the model's capability to capture true crack patches, resulting in an impressive F1 score of 99.2. InceptionV3 delivered a respectable accuracy of 97.3% and a precision of 94.3%, but it achieved a recall of 1, indicating that it may overlook certain crack patches. The F1 score, at 97.06, represents a balanced measure of precision and recall. EfficientNetV2 excelled with an accuracy of 99.6% and a high precision of 99.3%, which means it effectively minimized false positives. The recall of 1 suggests its ability to capture all true crack patches. The F1 score of 99.6 demonstrates an exceptional balance between precision and recall, making it a top-performing model in this evaluation.

For the metric of accuracy, EfficientNetV2 obtained the best result with an accuracy of 99.6%, while VGG19 achieved the lowest accuracy at 92.2%. In terms of precision, EfficientNetV2 demonstrated the best performance with a precision score of 99.3%, while Inception V3 had the lowest precision at 94.3%. Regarding recall, models except VGG19 stood out as the top performer with a recall score of 100%. For the F1 score, EfficientNetV2 led the way with an F1 score of 99.6, reflecting its excellent balance between precision and recall. In contrast, Vgg19 had the least favorable F1 score at 92.9. In conclusion, our extensive evaluation of concrete crack detection models revealed notable variations in their performance across multiple metrics. Among the models considered, EfficientNetV2 emerged as the standout performer, consistently achieving the highest accuracy, precision, and recall scores, resulting in an exceptional F1 score of 99.6%.

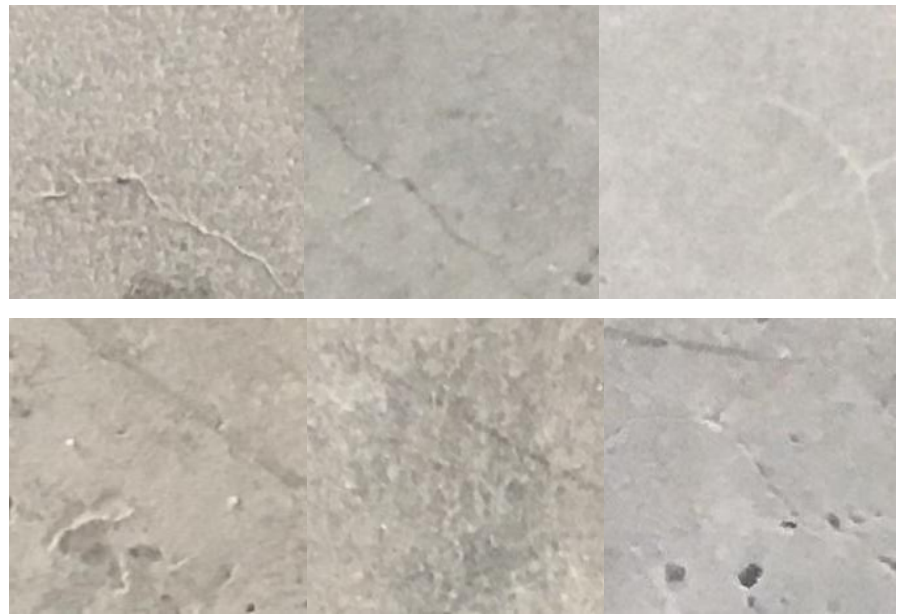

**Fig. 3. Some false positives by proposed models**

**Table 1. Deep Learning Models: Number of layers and parameters.**

| Model | Number of Convolutional Layers | Number of Parameters (Millions) |
|---|---|---|
| VGG19 | 19 | 143 |
| ResNet50 | 50 | 23 |
| InceptionV3 | 48 | 21 |
| EfficeinNetV2 | 237 (B0) | 25 |

**Table 2. Comparison of the different models: the pre-trained models (without training)**

| Model | Accuracy | Precision | Recall | $F1_{score}$ |
|---|---|---|---|---|
| VGG19 | 0.892±0.0022 | 0.947±0.0020 | 0.840±0.0039 | 0.900±0.00055 |
| ResNet50 | 0.975±0.0032 | 0.957±0.0022 | 0.975±0.0047 | 0.985±0.0031 |
| InceptionV3 | 0.942±0.0069 | 0.925±0.033 | 0.998±0.075 | 0.961±0.0055 |
| EfficientNetV2 | 0.989±0.0039 | 0.989±0.00098 | 0.997±0.0025 | 0.989±0.0039 |

**Table 3. Comparison of the different models: fine tuning the pre-trained models.**

| Model | Accuracy | Precision | Recall | $F1_{score}$ |
|---|---|---|---|---|
| VGG19 | 0.922±0.0028 | 0.974±0.0022 | 0.883±0.00084 | 0.929±0.00051 |
| ResNet50 | 0.994±0.0019 | 0.988±0.0015 | 0.999±0.0052 | 0.992±0.0028 |
| InceptionV3 | 0.973±0.016 | 0.943±0.038 | 0.999±0.085 | 0.970±0.0064 |
| EfficientNetV2 | 0.996±0.0042 | 0.993±0.0079 | 0.999±0.0033 | 0.996±0.0059 |

Its ability to effectively minimize false positives while capturing all actual crack patches underscores its robustness in this specific task. On the other hand, VGG19 exhibited the least favorable results in our assessment, particularly in terms of accuracy and recall, indicating a potential weakness in identifying certain crack patches. Overall, our study underscores the importance of selecting the right deep learning model for the specific demands of concrete crack detection, and EfficientNetV2 stands out as a strong contender in this context, showcasing its efficiency and accuracy in this critical application.

*4.4 Statistically Analysis*

To determine whether there is a statistically significant difference between the models in terms of classification performance metrics (Accuracy, Precision, Recall, F1 Score), we can use statistical tests such as Analysis of Variance (ANOVA) for comparing multiple groups. ANOVA can help determine if there are significant differences between the means of the different models. The p-values provided by the test will indicate whether there are significant differences between the models for each metric. If p-values are less than your chosen significance level (e.g., 0.05), we can conclude that there are significant differences.

Based on the provided p-values from the ANOVA analysis, it appears that there are significant differences between the models for all four-performance metrics: Accuracy, Precision, Recall, and F1 Score. For accuracy, the p-value is very close to zero (2.4e-09), which indicates a significant difference between the models in terms of accuracy. The p-value of precision is also very close to zero (6.9e-10), suggesting a significant difference between the models in terms of precision. The p-value (1.8e-07) is still quite small, indicating a significant difference between the models in terms of recall. The p-value (1e-09) is very close to zero, indicating a significant difference between the models in terms of the F1 Score. In summary, the statistical analysis indicates that there are significant differences in performance between the models for all four metrics. These results suggest that the choice of model has a significant impact on classification performance.

## 5.  CONCLUSION

In this study, we conducted a rigorous evaluation of deep learning models fine-tuned for the task of crack detection in concrete surfaces. Four popular deep learning architectures were employed, and each model was equipped with a custom fully connected layer designed for binary classification. The performance of these models was assessed using various metrics, including accuracy, precision, recall, and F1 score. Our findings unveiled compelling insights into the capabilities of these models in tackling the concrete crack detection challenge. Among the models we examined, EfficientNetV2 emerged as the standout performer. Its consistent and remarkable results across all metrics underscore its efficacy and efficiency in the specific task of concrete crack detection. With an outstanding accuracy of 99.6% and a precision of 99.3%, EfficientNetV2 excelled in minimizing false positives, an essential aspect of this application. Additionally, it achieved a perfect recall score of 1, indicating its capability to capture all actual crack patches. The model's overall performance was further highlighted by an impressive F1 score of 99.6, demonstrating a harmonious balance between precision and recall.